\pdfoutput=1

\documentclass[11pt]{article}
\usepackage{color}
\usepackage{times}
\usepackage{epsfig}
\usepackage{graphicx}
\usepackage{amsmath}
\usepackage{amsthm}
\usepackage{amssymb}
\usepackage{mathtools}
\usepackage{multirow}
\usepackage{bbm}
\usepackage{dsfont}

\usepackage[table, dvipsnames]{xcolor}

\usepackage[utf8]{inputenc}
\usepackage{booktabs,siunitx}
\usepackage{multirow}
\usepackage{multicol}
\usepackage{amsmath}
\usepackage{subcaption}
\usepackage{caption}
\usepackage{graphicx}
\usepackage{pifont}

\usepackage{tikz}
\usetikzlibrary{shapes,arrows,patterns}
\usetikzlibrary{positioning}
\usetikzlibrary{calc}

\newcommand{\cut}[1]{}

\newcommand{\postspace}{\vskip -3mm}
\newcommand{\minipostspace}{\vskip -2mm}

\definecolor{red}{RGB}{255, 117, 115}
\definecolor{green}{RGB}{171, 255, 175}

\usepackage{ACL2023}

\usepackage{times}
\usepackage{latexsym}
\usepackage{multirow}
\usepackage{makecell}
\usepackage{tcolorbox}
\usepackage{enumitem}

\usepackage[T1]{fontenc}

\usepackage[utf8]{inputenc}

\usepackage{microtype}

\usepackage{inconsolata}

\definecolor{cadmiumgreen}{rgb}{0.0, 0.42, 0.24}
\definecolor{cardinal}{rgb}{0.77, 0.12, 0.23}
\definecolor{cadmiumred}{rgb}{0.89, 0.0, 0.13}

\newtcolorbox[list inside=prompt,auto counter,number within=section]{prompt}[1][]{
    fontupper=\ttfamily\footnotesize,
    boxsep=5pt,
    left=0pt,
    right=0pt,
    top=0pt,
    bottom=0pt,
    boxrule=1pt,
    #1,
}

\title{Distilling Large Language Models using Skill-Occupation Graph Context for HR-Related Tasks}

\newcommand\askip{\hspace{10pt}}

\author{\makecell{Pouya Pezeshkpour$^1$ \askip Hayate Iso$^1$ \askip Thom Lake$^2$\\Nikita Bhutani$^1$ \askip Estevam Hruschka$^1$}\\
  $^1$Megagon Labs \askip $^2$Indeed\\
  \texttt{\{pouya,hayate,nikita,estevam\}@megagon.ai \askip tlake@indeed.com} 
}

\begin{document}
\maketitle
\begin{abstract}
    Numerous HR applications are centered around resumes and job descriptions. While they can benefit from advancements in NLP, particularly large language models, their real-world adoption faces challenges due to absence of comprehensive benchmarks for various HR tasks, and lack of smaller models with competitive capabilities. In this paper, we aim to bridge this gap by introducing the Resume-Job Description Benchmark (RJDB). We meticulously craft this benchmark to cater to a wide array of HR tasks, including matching and explaining resumes to job descriptions, extracting skills and experiences from resumes, and editing resumes. To create this benchmark, we propose to distill domain-specific knowledge from a large language model (LLM). We rely on a curated skill-occupation graph to ensure diversity and provide context for LLMs generation. Our benchmark includes over 50 thousand triples of job descriptions, matched resumes and unmatched resumes. Using RJDB, we train multiple smaller student models. Our experiments reveal that the student models achieve near/better performance than the teacher model (GPT-4), affirming the effectiveness of the benchmark. Additionally, we explore the utility of RJDB on out-of-distribution data for skill extraction and resume-job description matching, in zero-shot and weak supervision manner. We release our datasets and code\footnote{\url{https://github.com/megagonlabs/rjdb}} to foster further research and industry applications.
%
\end{abstract}

\section{Introduction}
In organizational recruitment, resumes and job descriptions play a pivotal role, facilitating identification of potential candidates and informing hiring decisions \citep{zimmermann2016data, guo2021smarter, ali2022resume}. Natural Language Processing (NLP) algorithms play a crucial role in enhancing this process, unraveling valuable information and insights embedded within resumes and job descriptions. This include information such as job titles, skill sets, work history, and educational background. Leveraging such information can facilitate a broad variety of HR tasks such as aligning candidates with job openings, streamlining the resume screening process, and aiding in salary negotiation. Deploying NLP models for such use cases requires a diverse and representative dataset covering a wide range of resumes and job descriptions. However, there is a noticeable absence of publicly available large-scale benchmarks tailored for this purpose. Additionally, many companies heavily rely on proprietary, in-house data to train NLP models. However, such datasets often carry inherent biases and noise, and only provide necessary annotation for limited downstream tasks.

\begin{figure}[t]
    \centering
    \includegraphics[width=\columnwidth]{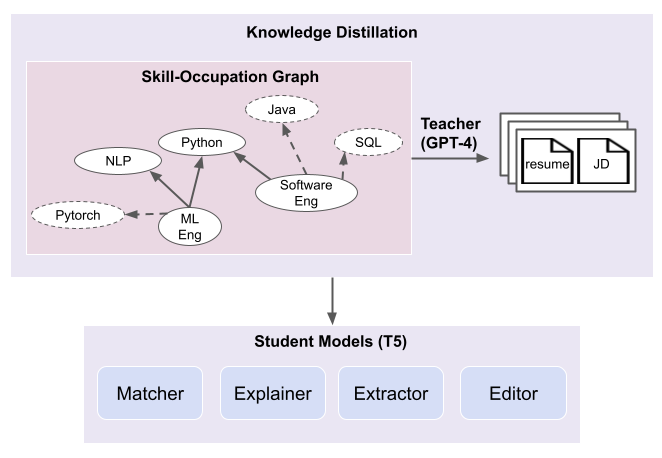}
    \caption{Creating a benchmark for HR-related tasks via large language model distillation. We start by meticulously sampling subgraphs from our curated skill-occupation graph. Then, we utilize the skills and occupations contained within these subgraphs as contextual guidance to steer GPT-4 in the generation of resumes and job descriptions encompassing a variety of tasks.}
    \label{fig:over}
\end{figure}
 
In recent years, several attempts have been made to create/augment datasets by distilling knowledge from existing large language models (LLMs) \citep{kim2022soda, gu2023knowledge,li2023symbolic,gu2023knowledge}. The objective is to extract the knowledge stored in a large teacher model such as GPT-4 \citep{openai2023gpt-4}, into a benchmark for a specific task. Subsequently, this benchmark becomes a valuable resource for training smaller student models, allowing them to emulate the performance of the teacher model in the targeted task. These  student models present a practical solution for real-world applications, owing to their reduced size. 

Building on the success of prior work in knowledge distillation from LLMs, we propose to distill knowledge from LLMs to generate a multi-task resume-job description benchmark. 
As large language models struggle to generate high quality and accurate output in low resource domains \citep{bang2023multitask}, such as HR, they require to be provided with necessary guidance as the context \citep{agrawal2022large,singhal2022large,jin2023genegpt}. Skills and past experiences (occupations) are recognized as the foundational information required to solve various HR-related tasks \citep{qin2020enhanced, fu2021incongruent, sun2021market}. Consequently, to guide the knowledge distillation effort from LLMs, using skill-occupation graph---a bipartite graph connecting occupations with their required skills---as the context, emerges as a natural candidate. 
Since, to the best of our knowledge, there is no diverse publicly available skill-occupation graph, we curate our own graph in two steps: Firstly, we initiate the graph by harvesting data pertaining to technology-related occupations and their skills from Dice\footnote{\url{https://www.dice.com}. We extracted the data from \url{https://www.kaggle.com/datasets/PromptCloudHQ/us-technology-jobs-on-dicecom}}. Subsequently, we extend this graph by first extracting general occupations from the US Bureau of Labor 
Statistics\footnote{\url{https://www.bls.gov/oes/current/oes_stru.htm}}, and then leveraging GPT-4 to generate required skills for these occupations. 

Leveraging the structure of the graph, we employ a sampling process to assemble a diverse set of interconnected skills and occupations. Then, we iteratively provide sampled subgraphs as context to GPT-4 to generate more than 50,000 triples, each comprising a job description, a matched resume, and an unmatched resume. We craft the generation pipeline such that it yields triples that serve as resources for training and evaluating models across several pivotal tasks: 
(1) \textbf{Job-Candidate Matching:} Assessing the compatibility between job openings and potential candidates.
(2) \textbf{Counterfactual Explanation for Matching:} Exploring explanations and reasoning behind job-candidate matches.
(3) \textbf{Information Extraction:} Extracting pertinent details, such as skills and occupations, from both resumes and job descriptions.
And, (4) \textbf{Resume Editing:} Facilitating the enhancement and refinement of resumes for candidates. We provide an overview of our pipeline to generate RJDB in Figure \ref{fig:over}. 

After creating the resume-job description benchmark (RJDB), we train individual student models based on Flan-T5 base \citep{chung2022scaling} for each specific task within the dataset. Our student models demonstrate comparable/superior performance across different tasks, when compared against the GPT-4 teacher model over held-out data from the benchmark. To explore the broader utility of RJDB, we extend our investigation to include out-of-distribution data, specifically focusing on two tasks: resume-job matching and skill extraction. Remarkably, our student models, fine-tuned on RJDB, excel in these tasks. We also find that the models fine-tuned on out-of-distribution data when further fine-tuned using RJDB, yield enhanced performance and improved generalization. 

\noindent\textbf{Contributions} The main contributions of this paper are as follows: (1) We create a multi-task resume-job description benchmark (RJDB) to overcome existing datasets limitations in the HR domain. (2) We curated a diverse skill-occupation graph and utilized it as the context guiding 
our benchmark generation process. (3) We introduced a novel distillation pipeline for HR; our approach comprises various innovative components that leverage domain-specific expertise and harness the world-knowledge from GPT-4 to effectively address numerous existing challenges in the HR domain. (4) Training multiple student models using RJDB on the proposed tasks, we release strong baselines that achieved performance on par with or surpassing the teacher model (GPT-4). Finally, (5) we explored potential use cases of RJDB in out-of-distribution data.
We have made our data, models, and code publicly available, aiming to facilitate and advance progress within the HR domain.

\section{Distilling LLMs For HR-Related Tasks}

To generate high-quality documents in HR domain, it is crucial to provide knowledge intensive context to guide large language models. Furthermore, since our goal is to generate documents satisfying multiple HR-related downstream tasks, this context should contain necessary annotations for these tasks. As a result, subgraphs from a skill-occupation graph naturally emerge as a suitable candidate to serve as context for guiding LLM generation. 
Leveraging a skill-occupation graph to distill knowledge from LLMs presents us with three distinct challenges: (1) The scarcity of publicly available skill-occupation graphs that cover a wide range of general occupations. (2) The need to sample subgraphs that not only provide the necessary guidance for LLMs but also yield documents that align with real-world distributions. And, (3) the requirement to generate documents that are both diverse and faithful to the provided context while offering essential annotations for downstream tasks. 
In this section, we take on these challenges by first curating our skill-occupation graph and then introducing our generation pipeline to construct our multi-task benchmark. Following the introduction of our pipeline, we then proceed to conduct a comprehensive quality assessment of the generated documents.

\subsection{Skill-Occupation Graph}

The skill-occupation graph \citep{dave2018combined,de2021job, bovskoski2022occupation} is a powerful tool in the realm of workforce development and career matching, defined as a bipartite graph that links occupations with their required skills. 
Our objective is to extract subsets from this graph to guide the generation of resumes and job descriptions. 
To the best of our knowledge, the only publicly accessible skill-occupation graph is sourced from DICE which only covers technology related occupations.  
As a result, to generate diverse resumes and job descriptions, we need to construct our own graph encompassing a broader variety of occupational categories. 
Beyond data availability, the most significant challenge in developing such a graph is ensuring its representation aligns with real-world distribution. 
Thus, in this section, as a proactive step in overcoming these challenges, we start constructing our graph with a foundation in technology-related graph from DICE. Building upon this foundation, we extend the graph by extracting general occupations from the US Bureau of Labor Statistics. We then generate required skills for each one of those occupations by prompting GPT-4.

\paragraph{Generating Skills for General Occupations}
To diversify our graph beyond technology-related occupations, we incorporate 1,112 diverse occupations sourced from the US Bureau of Labor Statistics. The process of generating the required skills for these occupations involves several steps. Initially, we match each occupation with the closest counterpart in 
our in-house proprietary skill-occupation graph, leveraging Phrase-BERT \citep{wang2021phrase}. Subsequently, we extract the number of provided skills associated with the matched occupation from the graph. This process is vital for ensuring that distribution of skills in the generated graph aligns with real-world distribution. For each occupation and its respective extracted number of skills, denoted as $n$, we employ GPT-4 to generate required skills for the occupation by using the prompt:
\begin{prompt}[title={\footnotesize\texttt{Prompt \thetcbcounter: Prompt for skill generation}}, label=prompt:skill_gen]
    Generate \{n\} number of required skills necessary for the occupation \{OCCUPATION\}.
\end{prompt}

\paragraph{Filtering the Curated Graph}
In order to sample diverse subgraphs from our curated graph while effectively filtering out rare skills and occupations, we employ a clustering strategy. We adopt density-based clustering and initiate the process by selecting a random occupation as a seed, then gradually expand the cluster to encompass any node within a two-hop radius. This expansion continues iteratively until all occupations are included within a cluster. Subsequently, we eliminate clusters that contain fewer than 10 occupational nodes. As a result of this process, we successfully partition and filter the graph into 404 distinct clusters, with an average of approximately 84.2 occupational nodes within each cluster. A statistical overview of the resulting graph is presented in Table \ref{tab:sk-oc}.

\begin{table}[t!]
\small
\centering
\begin{tabular}{rrrrr}
\toprule 
\multirow{2}{*}{\bf\#Occ}&\multirow{2}{*}{\bf\#Skill}& \multirow{2}{*}{\bf\#Edges} & \bf\#Avg Skill& \bf\#Avg Occ\\
&& &\bf per Occ&\bf per Skill\\
\midrule
8275&14807&70661&8.5&4.8\\
\bottomrule
\end{tabular}
\caption{Data statistics of the curated skill-occupation graph.}
\label{tab:sk-oc}
\end{table}
\subsection{Generation Pipeline}


In the process of generating resumes and job descriptions, our approach involves sampling subgraphs from the curated skill-occupation graph to serve as context for LLMs. These subgraphs comprise sets of skills and past experiences (occupations) that are to be integrated into a candidate's resume and a job description. To ensure the quality and authenticity of generated documents, several necessary requirements must be met. 
Firstly, the number of sampled skills and experiences should align with the real-world distribution of these components within a resume or job description. Furthermore, the provided experiences should maintain a logical chronological order and incorporate temporal information. Beyond these prerequisites, we aim to create unbiased and diverse documents while simultaneously providing essential annotations for downstream tasks.
In this section, we shape our document generation pipeline to adhere to these requirements. 
Given the absence of readily available data to help addressing these requirements, we mostly rely on the world knowledge embedded in GPT-4 to fulfill them. 

\paragraph{Subgraph Sampling}
To initiate the subgraph sampling process, we begin by randomly selecting a cluster and an occupational node within that cluster as our starting node. To sample a subgraph, we implement random walks from the starting node and configure the random walks parameters to ensure that the expected number of sampled skills and experiences (occupations) align with predefined values. To specify the predefined number of experiences, in each sampling step, we randomly select a value ranging from 1 to 5. For the predefined number of skills, we turn to GPT-4 and pose the question:
\begin{prompt}[title={\footnotesize\texttt{Prompt \thetcbcounter: Prompt for subgraph sampling}}, label=prompt:sampling]
    On average how many skills does a person with a job title of `\{Starting Node\}' may have listed in his or her resume?
\end{prompt}
This approach enables us to align the number of skills in generated documents with the real-world distribution, accomplished by using GPT-4's extensive world knowledge, tailored to a specific occupation.

\paragraph{Chronological Order of Experiences in a Sampled subgraph}
After sampling a subgraph, the past experiences are presented in a random sequence. However, in actual real-world resumes, past experiences typically follow a logical chronological order, reflecting an applicant's career progression over time. For instance, an applicant commonly starts as a Software Engineer before advancing to the role of a Senior Software Engineer. To rectify this, we turn to GPT-4's world knowledge by prompting it with the prompt:
\begin{prompt}[title={\footnotesize\texttt{Prompt \thetcbcounter: Prompt for ordering experiences}}, label=prompt:ordering]
    Given the previous experiences of individuals with \{list of Past Experiences\}, please arrange them in a chronological order based on the likelihood of encountering these experiences from earlier to later over time.
\end{prompt}
This aids us in establishing the correct chronological order for the sampled past experiences.

\begin{figure*}[t]
    \centering
    \includegraphics[width=0.9\textwidth]{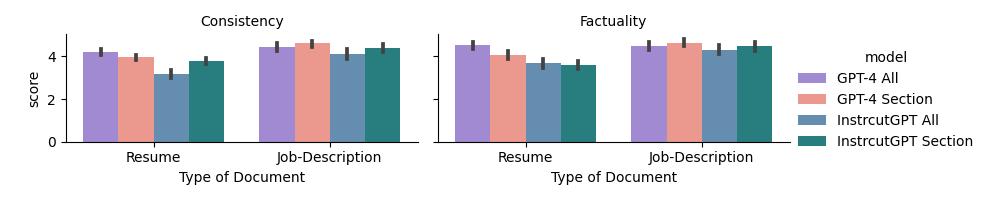}
    \caption{Evaluating the quality of generated documents using various approaches with G-eval.}
    \label{fig:geval}
    \postspace
\end{figure*}
\paragraph{Incorporating Temporal Information into Experiences}
We exclusively attribute temporal data to the past experiences within resumes and job descriptions. To accomplish this, we follow a process in which we randomly select a value between 1 and 5, signifying the number of years a particular experience spans. However, when generating resumes, rather than specifying years of involvement, we introduce a different approach. By selecting a random year between 2015 and 2023 (representing the last year of being active), we gradually reduce the number of years attributed to each experience. This method allows us to establish a specific time frame for each past experience, for instance, spanning from 2017 to 2021.

\paragraph{Diversifying Resumes based on Gender and Race}
In our pursuit of generating diverse resumes with respect to race and gender, we concentrate on the applicant's first name. Our approach involves extracting a list of the most popular given names that are uniformly distributed across various racial categories, sourced from Wikipedia \footnote{\url{https://en.wikipedia.org/wiki/List_of_most_popular_given_names}}. This results in a collection of approximately 700 first names, separately for both males and females. In each generation step, we prioritize maintaining a balanced distribution of male and female names across various occupation types, ensuring that every cluster comprises an equal number of generated resumes for both genders. Additionally, as these first names are uniformly distributed across different racial groups, uniformly sampling them further ensures the race diversity of generated resumes.

\paragraph{Task Specific Annotation}
As our objective is to construct a benchmark capable of encompassing multiple tasks, we must ensure that our generation process aligns with the distinct requirements of each task. Given that our approach generates both resumes and job descriptions from the same initial set of skills and experiences, the prerequisites for both the \textbf{matching} and \textbf{extracting} tasks are inherently fulfilled. To facilitate the necessary annotations for \textbf{resume editing} and \textbf{matching explanation}, alongside generating a job description and a matched resume, we introduce an unmatched resume by deliberately altering certain elements within the matched resume. Specifically, we randomly remove between 1 to 5 skills and decrease the duration of engagement in the last experience by randomly selecting a number equal to or less than the original years of involvement.

\subsection{Quality Assessment}
With our generation pipeline outlined, our next step involves determining whether we can leverage InstructGPT (text-davinci-003) \citep{ouyang2022training} or if a more advanced model like GPT-4 \citep{openai2023gpt-4} is required for resume-job description distillation. Additionally, we aim to ascertain whether it's more advantageous to generate documents section by section or produce the entire document in one go. To address these inquiries, we delve into the evaluation of \textit{consistency} and \textit{factuality} in the generated documents, employing G-eval \citep{liu2023gpteval}, an evaluation method based on large language models. Also, we conduct a human study to gain a deeper understanding of the quality of generated documents.

\paragraph{LLM-based Evaluation}
We employ G-eval to conduct automated assessments of the \textit{consistency} and \textit{factuality} of the generated documents, to explore the impact of different LLMs and generation methods (section by section or in its entirety). G-eval evaluation only requires the input document, along with criteria and chain-of-thought style \citep{wei2022chain} instructions for scoring the document on a scale of 1 to 5 (detailed prompts are available in Appendix). The outcomes of our evaluation are presented in Figure \ref{fig:geval}. The results indicate that, in general, GPT-4's approach of generating the entire document at once demonstrates superior performance across both criteria. As a result, we consider this approach as the primary component in our generation pipeline in the remainder of this paper.

\begin{table}[t!]
\small
\centering
\begin{tabular}{rrr}
\toprule 
&\bf Factulity&\bf Consistency\\
\midrule
Dice& 63.4&62.7\\
Generated&\bf 67.9& \bf 65.4\\
\bottomrule
\end{tabular}
\caption{User study on the quality of generated job-descriptions.}
\label{tab:user}
\postspace
\end{table}

\paragraph{Human-based Evaluation}
We further assess the quality of generated documents through a user study. Since resumes can contain sensitive data, we only focus on the quality of generated job descriptions and compare it to real job descriptions gathered from Dice\footnote{\url{https://www.dice.com}}. We evaluate the quality of generated versus real world job description through two criteria: (1) \textbf{Realisticness}, measuring how probable is the provided job description to be a human written job description for an actual job. 
(2) \textbf{Consistency}, measuring how well the various elements within the job description, including job title, skills, and requirements, align with one another. We consider 100 generated job descriptions and extract 100 most similar job descriptions from Dice using phrase-BERT \citep{wang2021phrase} on job titles. We provide each sample to 3 annotators asking them to score the document based on both criteria on as scale from 1-100. Then, we calculate the average score for each sample. 

The outcomes of the user study are presented in Table \ref{tab:user}. The results indicate that the generated job descriptions exhibit a higher degree of realisticness and consistency compared to real job descriptions. One possible explanation for the higher scores of generated job descriptions could be attributed to the fact that we consistently generate fixed sections for all job descriptions, whereas many of the Dice job descriptions lack well-defined sections.

\begin{table}[t]
\small
\centering
\begin{tabular}{rrrrr}
\toprule 
&\bf\#Doc&\bf Avg \#W&\bf Min \#W& \bf Max \#W\\
\midrule
JD&52K& 181.1& 70& 427\\
R-M&52K&101.9&30&604\\
R-U&52K&87.1&23&395\\
\bottomrule
\end{tabular}
\caption{Data statistics of RJDB. We report the number of generated documents (\#Doc), and also average, minimum, and maximum number of words (\#W) for job descriptions (JD), matched resumes (R-M), and unmatched resumes (R-U).}
\label{tab:rjdb}
\minipostspace
\end{table}
\begin{table}[t]
\small
\centering
\begin{tabular}{rrrrr}
\toprule 
&\bf Avg \#Skills&\bf Avg \#Exp\\
\midrule
Sampled&6.43&2.44\\
Removed&1.9&-\\
\bottomrule
\end{tabular}
\caption{The average number of skills and past experiences in each triple, as well as the average number of skills removed to generate the unmatched resume from the matched one.}
\label{tab:ext}
\postspace
\end{table}
\begin{figure}[t]
    \centering
    \includegraphics[width=0.9\columnwidth]{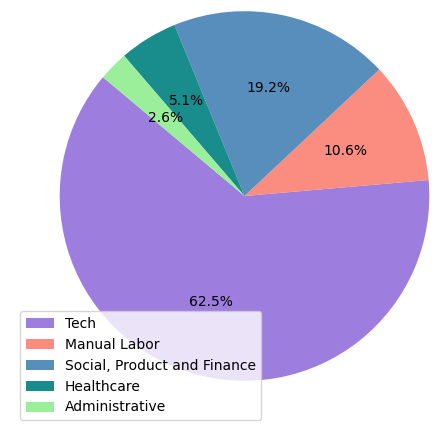}
    \caption{Distribution of different job categories over generated documents.}
    \label{fig:cat}
\postspace
\end{figure}

\section{Resume-Job Description Benchmark (RJDB)}
Incorporating our generation pipeline, our goal was to create the Resume-Job Description Benchmark (RJDB), which comprises 52,000 triples of job descriptions, matched resumes, and unmatched resumes. Details regarding the prompt used for generating these triples can be found in Appendix. For insights into the data statistics, including the number of tokens in the generated documents, please refer to Tables \ref{tab:rjdb}. Additionally, Tables \ref{tab:ext} present the average number of sampled skills, experiences, and the average number of removed skills to create the unmatched resume from the matched ones.

\begin{figure*}[t]
    \centering
        \captionsetup[subfigure]{justification=centering}
        \begin{subfigure}{.32\linewidth}
            \centering
            \includegraphics[width=\textwidth]{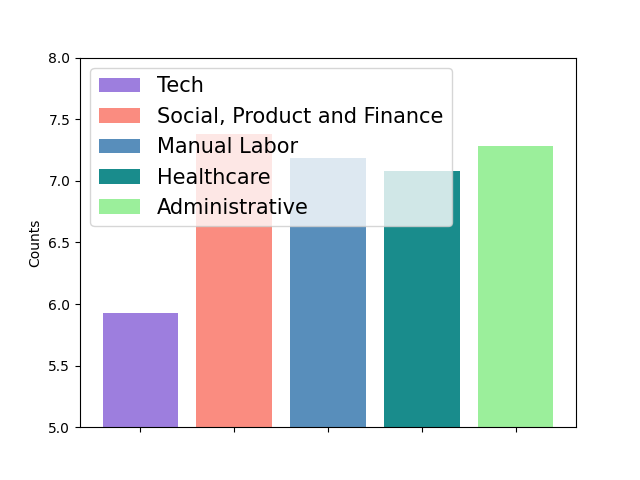}
            \caption{Distribution of average number of skills.}
            \label{fig:as}
        \end{subfigure}
        \begin{subfigure}{.32\linewidth}
            \centering
            \includegraphics[width=\textwidth]{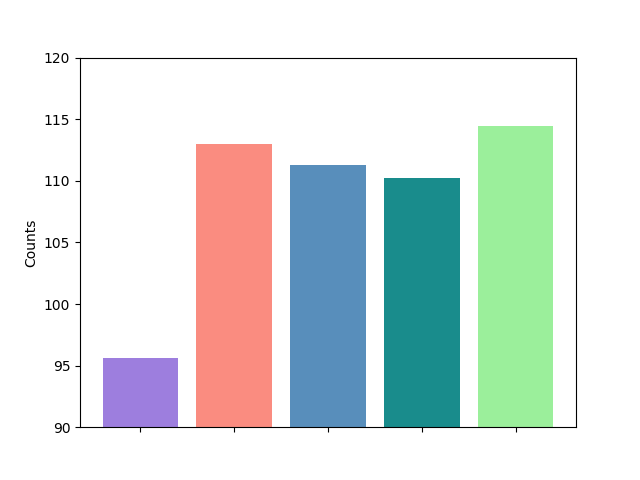}
            \caption{Distribution of average number of words per resume.}
            \label{fig:tr}
        \end{subfigure}
        \begin{subfigure}{.32\linewidth}
            \centering
            \includegraphics[width=\textwidth]{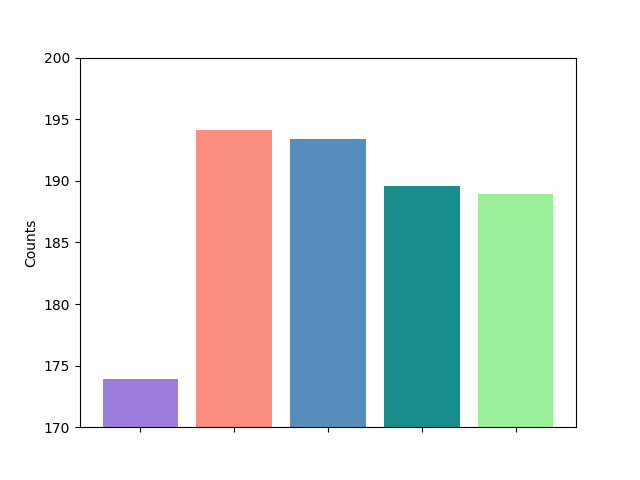}
            \caption{Distribution of average number of words per job description.}
            \label{fig:tj}
        \end{subfigure}
    \caption{Distribution of number of skills and words in generated documents.}
    \label{fig:finer}
    \postspace
\end{figure*}

To delve deeper into the distribution of the generated documents, we categorized them into five distinct groups: (1) tech, (2) social, product, and finance, (3) manual labor, (4) healthcare, and (5) administrative. The distribution of triples across these categories is depicted in Figure \ref{fig:cat}. Furthermore, a more detailed breakdown of the average number of skills, and words in each document can be found in Figure \ref{fig:finer}. Remarkably, documents related to the tech industry exhibit a lower average number of skills and words. 
This phenomenon may be attributed to the distinctive graph structure surrounding tech-related occupations, as well as the diversity of these occupations in the graph covering different level of expertise.

\section{Student Models}
Now that we have successfully generated the RJDB, the next crucial step involves the training of FLAN-T5 base student models (which we refer to as T5 in the remainder of paper). This includes training T5 on matching, explaining, extracting, and editing tasks. Upon fine-tuning T5 on these tasks we compare it against our teacher model, GPT-4. This comparative analysis is essential to measure the capabilities and potential of student models in capturing knowledge of teacher model. To achieve this, we divide the RJDB to train, test, and dev set with the size of 50000, 1000, and 1000 samples respectively.  

\paragraph{Matcher and Explainer} 
We fine-tune the T5 student model for both matching and explaining tasks simultaneously. When presented with pairs of resumes and job descriptions, our goal here is not only to predict their compatibility but also to explain the reasons behind this determination by providing a set of matching or mismatching skills and experiences. For negative pairs, we consider modifications on the matched resume, to create the un-matched version, to be the explanation, while for positive examples we consider the set of seed skills and experiences as the explanation. Considering the complicated nature of evaluating explanations, here, we only report the percentage of modifications that appear in the explanations in negative pairs, to evaluate the explaining capability of models. The results for matching and explaining capability of models are presented in Table \ref{tab:match}. Remarkably, the student model exhibit an outstanding level of performance, outperforming GPT-4 (teacher) in both tasks. Based on T5 performance, it seems that if provided with a big enough benchmark a smaller model can achieve comparable performance with a much larger teacher model on matching and explaining tasks when tested over the data from the same distribution as training data.  

\paragraph{Extractor 
}
We also fine-tune a student model to extract skills and past experiences from resumes and job descriptions. The results of our student models performances are presented in Table \ref{tab:extract}. Both models generally exhibit superior performance when extracting information from resumes compared to job descriptions. This observation may be attributed to the typically shorter length of resumes, and possibly more explicit appearance of information in resumes compared to job descriptions. 
Furthermore, in the extraction of past experiences, student models consistently either outperform or perform at a comparable level to the teacher model in both resume and job descriptions. Conversely, in the task of skill extraction, while student models exhibit similar average F1 performance, a noticeable disparity in accuracy becomes apparent between the teacher and the student model. Suggesting that this task may demand a higher level of reasoning, which emphasizes the potential need for a larger student model or a more complex training procedure.

\paragraph{Editor} 
Our final student model is tailored for the resume editing task, focusing on the reconstruction of the skills and experiences section within the unmatched resume to create the matched version. We use the entirety of the unmatched resume as an input and incorporating the modifications made to the matched resume to create the unmatched counterpart as an extra signal, which includes the skills that were removed and changes to the last experience. In evaluating the model's performance, in addition to ROUGE2 score \citep{lin2004rouge}, we employ $F_{add}$~\citep{xu2016optimizing}, measuring F1 score by rewarding the n-grams (bi-grams in our setting) that appear in the model output and the reference document (extra signals) but did not appear in the input document (unmatched resume). The outcomes of models' performance are provided in Table \ref{tab:edit}. Both models show higher level of performance when editing experience section. 
Additionally, T5 excels in the $F_{add}$ metric while underperforming in the ROUGE metric comparing to GPT-4, highlighting the observation that student models tend to produce more concise and precise contents.

\begin{table}[t]
\small
\centering
\begin{tabular}{lrrr}
\toprule 
\multirow{2}{*}{\bf Models} & \multicolumn{2}{c}{\bf Matching}&  \bf Explaining\\
\cmidrule(lr){2-3}
\cmidrule(lr){4-4}
& Acc& F1 & Acc\\
\midrule
GPT-4&83.2&82.8&37.8\\
T5&\bf 93.1&\bf 93.1&\bf 53.7\\
\midrule
\end{tabular}
\caption{Performance of the teacher and student models on matching and explaining tasks.}
\label{tab:match}
\minipostspace
\end{table}

\begin{table}[t]
\small
\centering
\begin{tabular}{llrrrr}
\toprule 
&\multirow{2}{*}{\bf Models} & \multicolumn{2}{c}{\bf Skill}&  \multicolumn{2}{c}{\bf Experience}\\
 \cmidrule(lr){3-4}
\cmidrule(lr){5-6}
&& Acc& F1 & Acc& F1\\
\midrule
\multirow{2}{*}{\rotatebox[origin=c]{90}{\bf Res}}&GPT-4&\bf 79.2&\bf81.9&\bf 56.6&57.7\\
&T5&60.8&73.9&55.5&\bf 64.5\\
\midrule
\multirow{2}{*}{\rotatebox[origin=c]{90}{\bf JD}}&GPT-4&\bf 83.9&\bf 77.1&41.5&40.1\\
&T5&60.5&73.6&\bf 53.3&\bf 63.1\\
\midrule
\end{tabular}
\caption{Performance of the teacher and student models on information extraction task.}
\label{tab:extract}
\postspace
\end{table}

\begin{table}[t]
\small
\centering
\begin{tabular}{llrrrr}
\toprule 
&\bf Models& \bf ROUGE & $\bf{F_{add}}$\\
\midrule
\multirow{2}{*}{\rotatebox[origin=c]{90}{\bf Skill}}&GPT-4&\bf 0.347&0.262\\
&T5&0.278&\bf 0.271\\
\midrule
\multirow{2}{*}{\rotatebox[origin=c]{90}{\bf Exp}}&GPT-4&\bf 0.391&0.374\\
&T5&0.328&\bf 0.513\\
\midrule
\end{tabular}
\caption{Performance of the teacher and student models on resume editing task.}
\label{tab:edit}
\postspace
\end{table}

\section{Out-of-Distribution Use Cases}
To explore the broader applications of RJDB, we delve into scenarios focusing on out-of-distribution cases. This encompasses two key aspects: zero-shot adoption, where we apply our models fine-tuned on RJDB to outside data, and weak supervision, which involves using our dataset to further fine-tune models trained on out-of-distribution data. Since, the resume editing and explanation tasks remain relatively under-explored, and finding available resources for them proved to be a challenge, here, we only focus on matching and extraction tasks. For matching, we utilize the Machop dataset created by \citet{wang2022machop} from Indeed data, containing 2000 resume and job-description pairs divided into 1200, 400, 400 samples as train, test, and dev set respectively. Additionally, we use 688 resumes and their annotated skills provided in Machop for skill extraction task. 

\paragraph{Zero-Shot Adaptation in Matching}
For adopting our student model to be able to zero-shot predicting out-of-distribution data, in matching task, we need to incorporate noise into the data they are trained on. To do so, we consider un-matched generated resumes with maximum changes as positive samples. Also, we sample some random pairs of resume-job description for negative samples, creating a training set with around 12000 samples. As matching baselines, we consider Machop-Sequence \citep{wang2022machop} and Ditto \citep{li2020deep} which are both language model based solutions. The zero-shot performance of our student model is provided in Table~\ref{tab:match-out} (T5-R (Zero-shot)). 
Despite the fact that our T5-R (Zero-shot) was not trained on Machop data, surprisingly, not only it performs similarly to previously reported state-of-the-art model (Machop-S), it achieves higher performance in precision.

\paragraph{Weak Supervision in Matching}
HR-related tasks are highly dependent on social and economic conditions. For example, in an economic scenario where job applicants outnumber available positions, it may be necessary to implement a matching model that, while potentially sacrificing some true positive applicants for a job description, significantly reduces false positive cases. As a result, a continuous model adaptation becomes essential in order to address these ever-changing dynamics. RJDB can provide us with a controllable resource for tailoring existing models in a weak supervise manner to meet the requirements dictated by social and economic conditions. 
Fine-tuning a base Flan-T5 model on Machop dataset \citep{wang2022machop} - dataset (T5-M), the goal here is to explore the potential of utilizing RJDB data to further enhance this model. Considering the small training size of Machop dataset (1200 samples), fine-tuned T5 performs very poorly (Table \ref{tab:match-out}). 
We explore two scenarios: (1) one aimed at enhancing recall and (2) the other focused on improving precision. In the first scenario, we perform minimal additional fine-tuning on T5-M using a randomly selected subset of the created training set in previous part (which consists of 12,000 samples) containing 1,000 random samples, referred to as T5-M+R (Weak-Rec). In the second scenario, we randomly select 500 pairs of positive and negative resume-job description combinations from RJDB to further fine-tune T5-M (referred to as T5-M+R (Weak-Pre)). The results are presented in Table \ref{tab:match-out}. As the table illustrates, by leveraging RJDB, we can enhance recall and precision of T5-M by approximately 17.8\% and 8.2\%, respectively, with only marginal positive/negative effects on other metrics.

\begin{table}[t]
\small
\centering
\begin{tabular}{llrrr}
\toprule 
&&\bf F1 &\bf Rec &\bf Prec\\
\midrule
\multirow{3}{*}{\rotatebox[origin=c]{90}{\bf In}}&Ditto&66.4&81.7&56.0\\
&Machop-S&\bf 83.5&\bf 90.8&77.3\\
&T5-M&46.7&65.8&36.2\\
\midrule
\multirow{3}{*}{\rotatebox[origin=c]{90}{\bf Out}}&T5-R (Zero-shot)&80.5&77.5&\bf 83.8\\
&T5-M+R (Weak-Rec)&45.4&83.3&31.2\\
&T5-M+R (Weak-Pre)&50.0&56.7&44.4\\
\midrule
\end{tabular}
\caption{Matching task performances on out-of-distribution data.}
\label{tab:match-out}
\minipostspace
\end{table}
\begin{table}[t]
\small
\centering
\begin{tabular}{lrrrrrr}
\toprule 
\multirow{2}{*}{\bf Models} & \multicolumn{4}{c}{\bf Test-Machop}&  \multicolumn{2}{c}{\bf Human}\\
\cmidrule(lr){2-5}
\cmidrule(lr){6-7}
&\bf Acc&\bf F1 &\bf Rec &\bf Prec& \bf Acc&Avg \#\\
\midrule
T5-M&\bf 84.5&\bf 90.0&\bf 89.6&\bf 90.9&43.3& 0.18\\
T5-R&65.5&52.0&48.1&61.9&\bf 78.6&\bf 1.95\\
T5-M+R&83.1&88.1&87.7&89.3&55.0&0.35\\
\midrule
\end{tabular}
\caption{Skill extraction task on out-of-distribution.}
\label{tab:ex-out}
\postspace
\end{table}

\paragraph{Skill Extraction in Out-of-Distribution Data}
For skill extraction, we explore the effects of fine-tuning student models using RJDB on enhancing their generalization capabilities. To achieve this we adopt resumes provided in Machop, which are accompanied with their annotated skills from Indeed. 
We randomly select 100 resumes as the test set, 88 samples as dev set, and designate the remaining 500 resumes as the training set. We fine-tune a T5 model using three different datasets: (1) solely the Machop training set (T5-M), (2) solely a random sample of 1000 resumes from RJDB (T5-R), we choose a small set to be comparable to Machop training set, and (3) a combination of the Machop data and sampled resumes from RJDB (T5-M+R). The results of these models' performance on the Machop test set are presented in Table \ref{tab:ex-out}. Furthermore, to assess the generalization capabilities of these models, we manually verify the accuracy of additional skills extracted, i.e., predicted skills beyond the labeled ones, by each model from the test set resumes. The accuracy and average number of extra skills for each model are presented in Table \ref{tab:ex-out} (Human assessment). 
As demonstrated, T5-M+R can extract a significantly larger number of new skills with only a minor sacrifice in performance compared to T5-M. Similarly, while T5-R may underperform when compared to other models on the Machop test set, it successfully extracts a larger number of new skills with a higher accuracy. 

\section{Related Work}
Language models can be an invaluable asset for HR-related tasks, but their widespread adoption in real-world applications is hindered by the absence of publicly available multi-task datasets.

\paragraph{NLP for HR} In recent years, the integration of language models into human resources-related tasks has witnessed a notable surge in interest and innovation. These advanced natural language processing models have shown great promise in addressing the intricacies of HR tasks, ranging from resume parsing and job description matching to skill extraction and beyond. \citet{bian2020learning} proposed a multi-view co-teaching network for job-resume matching, designed to effectively leverage sparse and noisy interactions available between candidates and jobs. \citet{mahdi2021job, li2021method} introduced a BERT based \citep{devlin2019bert} information extraction from job descriptions and resumes respectively. Authors in \citet{fang2023recruitpro} proposed a skill-aware prompt learning module to improve the pre-trained model's adaptability to downstream HR-related tasks.

\paragraph{Knowledge Distillation from LLMs} Knowledge Distillation is a technique used for reducing the high computational demand of LLMs by transferring knowledge from a large teacher model to a smaller student one \citep{gu2023knowledge}. \citet{kim2022soda} curated a million-scale social dialogue dataset by distilling knowledge from InstructGPT. \citet{west2021symbolic} extracted commonsense symbolic knowledge from large language models, while \citet{li2023symbolic} introduced the distillation of chain-of-thought-style reasoning from LLMs to enhance the reasoning capabilities of smaller models.

\section{Conclusion}
We introduce the Resume-Job Description Benchmark (RJDB) by distilling knowledge from GPT-4. Starting from subgraphs sampled from our curated skill-occupation graph, we generate resume and job description catering to multiple HR-related tasks, from matching and explanation to skill and experience extraction. We generate 52,000 triples of job descriptions, matched and unmatched resumes and successfully train student models that rival or surpass the teacher model (GPT-4) on in-distribution data. Moreover, we extend our investigation to demonstrate the adaptability of RJDB in handling out-of-distribution data for skill extraction and resume-job description matching, using zero-shot and weak supervision techniques.
We believe RJDB lays a strong foundation for HR-related tasks, fostering the development of models and techniques that can bridge the gap between research and real-world applications in the field of HR.


\bibliography{main}

\begin{thebibliography}{32}
\expandafter\ifx\csname natexlab\endcsname\relax\def\natexlab#1{#1}\fi

\bibitem[{Agrawal et~al.(2022)Agrawal, Hegselmann, Lang, Kim, and
  Sontag}]{agrawal2022large}
Monica Agrawal, Stefan Hegselmann, Hunter Lang, Yoon Kim, and David Sontag.
  2022.
\newblock Large language models are few-shot clinical information extractors.
\newblock In \emph{Proceedings of the 2022 Conference on Empirical Methods in
  Natural Language Processing}, pages 1998--2022.

\bibitem[{Ali et~al.(2022)Ali, Mughal, Khand, Ahmed, and
  Mujtaba}]{ali2022resume}
Irfan Ali, Nimra Mughal, Zahid~Hussain Khand, Javed Ahmed, and Ghulam Mujtaba.
  2022.
\newblock Resume classification system using natural language processing and
  machine learning techniques.
\newblock \emph{Mehran University Research Journal Of Engineering \&
  Technology}, 41(1):65--79.

\bibitem[{Bang et~al.(2023)Bang, Cahyawijaya, Lee, Dai, Su, Wilie, Lovenia, Ji,
  Yu, Chung et~al.}]{bang2023multitask}
Yejin Bang, Samuel Cahyawijaya, Nayeon Lee, Wenliang Dai, Dan Su, Bryan Wilie,
  Holy Lovenia, Ziwei Ji, Tiezheng Yu, Willy Chung, et~al. 2023.
\newblock A multitask, multilingual, multimodal evaluation of chatgpt on
  reasoning, hallucination, and interactivity.
\newblock \emph{arXiv preprint arXiv:2302.04023}.

\bibitem[{Bian et~al.(2020)Bian, Chen, Zhao, Zhou, Hou, Song, Zhang, and
  Wen}]{bian2020learning}
Shuqing Bian, Xu~Chen, Wayne~Xin Zhao, Kun Zhou, Yupeng Hou, Yang Song, Tao
  Zhang, and Ji-Rong Wen. 2020.
\newblock Learning to match jobs with resumes from sparse interaction data
  using multi-view co-teaching network.
\newblock In \emph{Proceedings of the 29th ACM International Conference on
  Information \& Knowledge Management}, pages 65--74.

\bibitem[{Bo{\v{s}}koski et~al.(2022)Bo{\v{s}}koski, Perne, Redek, and
  Boshkoska}]{bovskoski2022occupation}
Pavle Bo{\v{s}}koski, Matija Perne, Tja{\v{s}}a Redek, and Biljana~Mileva
  Boshkoska. 2022.
\newblock Occupation similarity through bipartite graphs.
\newblock \emph{arXiv preprint arXiv:2202.11064}.

\bibitem[{Chung et~al.(2022)Chung, Hou, Longpre, Zoph, Tay, Fedus, Li, Wang,
  Dehghani, Brahma et~al.}]{chung2022scaling}
Hyung~Won Chung, Le~Hou, Shayne Longpre, Barret Zoph, Yi~Tay, William Fedus,
  Eric Li, Xuezhi Wang, Mostafa Dehghani, Siddhartha Brahma, et~al. 2022.
\newblock Scaling instruction-finetuned language models.
\newblock \emph{arXiv preprint arXiv:2210.11416}.

\bibitem[{Dave et~al.(2018)Dave, Zhang, Al~Hasan, AlJadda, and
  Korayem}]{dave2018combined}
Vachik~S Dave, Baichuan Zhang, Mohammad Al~Hasan, Khalifeh AlJadda, and
  Mohammed Korayem. 2018.
\newblock A combined representation learning approach for better job and skill
  recommendation.
\newblock In \emph{Proceedings of the 27th ACM International Conference on
  Information and Knowledge Management}, pages 1997--2005.

\bibitem[{de~Groot et~al.(2021)de~Groot, Schutte, and Graus}]{de2021job}
Maurits de~Groot, Jelle Schutte, and David Graus. 2021.
\newblock Job posting-enriched knowledge graph for skills-based matching.
\newblock \emph{arXiv preprint arXiv:2109.02554}.

\bibitem[{Devlin et~al.(2019)Devlin, Chang, Lee, and
  Toutanova}]{devlin2019bert}
Jacob Devlin, Ming-Wei Chang, Kenton Lee, and Kristina Toutanova. 2019.
\newblock Bert: Pre-training of deep bidirectional transformers for language
  understanding.
\newblock In \emph{Proceedings of the 2019 Conference of the North American
  Chapter of the Association for Computational Linguistics: Human Language
  Technologies, Volume 1 (Long and Short Papers)}, pages 4171--4186.

\bibitem[{Fang et~al.(2023)Fang, Qin, Zhang, Yao, Zhang, Zhu, Zhuang, and
  Xiong}]{fang2023recruitpro}
Chuyu Fang, Chuan Qin, Qi~Zhang, Kaichun Yao, Jingshuai Zhang, Hengshu Zhu,
  Fuzhen Zhuang, and Hui Xiong. 2023.
\newblock Recruitpro: A pretrained language model with skill-aware prompt
  learning for intelligent recruitment.
\newblock In \emph{Proceedings of the 29th ACM SIGKDD Conference on Knowledge
  Discovery and Data Mining}, pages 3991--4002.

\bibitem[{Fu et~al.(2021)Fu, Li, Feng, and Ye}]{fu2021incongruent}
Yan Fu, Nan Li, Juan Feng, and Qiang Ye. 2021.
\newblock Incongruent skills and experiences in online labor market.
\newblock \emph{Electronic Commerce Research and Applications}, 45:101025.

\bibitem[{Gu et~al.(2023)Gu, Dong, Wei, and Huang}]{gu2023knowledge}
Yuxian Gu, Li~Dong, Furu Wei, and Minlie Huang. 2023.
\newblock Knowledge distillation of large language models.
\newblock \emph{arXiv preprint arXiv:2306.08543}.

\bibitem[{Guo et~al.(2021)Guo, Gallagher, Sun, Tavoosi, and
  Min}]{guo2021smarter}
Feng Guo, Christopher~M Gallagher, Tianjun Sun, Saba Tavoosi, and Hanyi Min.
  2021.
\newblock Smarter people analytics with organizational text data:
  Demonstrations using classic and advanced nlp models.
\newblock \emph{Human Resource Management Journal}.

\bibitem[{Jin et~al.(2023)Jin, Yang, Chen, and Lu}]{jin2023genegpt}
Qiao Jin, Yifan Yang, Qingyu Chen, and Zhiyong Lu. 2023.
\newblock Genegpt: Augmenting large language models with domain tools for
  improved access to biomedical information.
\newblock \emph{ArXiv}.

\bibitem[{Kim et~al.(2022)Kim, Hessel, Jiang, Lu, Yu, Zhou, Bras, Alikhani,
  Kim, Sap et~al.}]{kim2022soda}
Hyunwoo Kim, Jack Hessel, Liwei Jiang, Ximing Lu, Youngjae Yu, Pei Zhou,
  Ronan~Le Bras, Malihe Alikhani, Gunhee Kim, Maarten Sap, et~al. 2022.
\newblock Soda: Million-scale dialogue distillation with social commonsense
  contextualization.
\newblock \emph{arXiv preprint arXiv:2212.10465}.

\bibitem[{Li et~al.(2023)Li, Hessel, Yu, Ren, Chang, and Choi}]{li2023symbolic}
Liunian~Harold Li, Jack Hessel, Youngjae Yu, Xiang Ren, Kai-Wei Chang, and
  Yejin Choi. 2023.
\newblock Symbolic chain-of-thought distillation: Small models can also" think"
  step-by-step.
\newblock \emph{arXiv preprint arXiv:2306.14050}.

\bibitem[{Li et~al.(2021)Li, Shu, Zhai, and Lin}]{li2021method}
XiaoWei Li, Hui Shu, Yi~Zhai, and ZhiQiang Lin. 2021.
\newblock A method for resume information extraction using bert-bilstm-crf.
\newblock In \emph{2021 IEEE 21st International Conference on Communication
  Technology (ICCT)}, pages 1437--1442. IEEE.

\bibitem[{Li et~al.(2020)Li, Li, Suhara, Doan, and Tan}]{li2020deep}
Yuliang Li, Jinfeng Li, Yoshihiko Suhara, AnHai Doan, and Wang-Chiew Tan. 2020.
\newblock Deep entity matching with pre-trained language models.
\newblock \emph{arXiv preprint arXiv:2004.00584}.

\bibitem[{Lin(2004)}]{lin2004rouge}
Chin-Yew Lin. 2004.
\newblock Rouge: A package for automatic evaluation of summaries.
\newblock In \emph{Text summarization branches out}, pages 74--81.

\bibitem[{Liu et~al.(2023)Liu, Iter, Xu, Wang, Xu, and Zhu}]{liu2023gpteval}
Yang Liu, Dan Iter, Yichong Xu, Shuohang Wang, Ruochen Xu, and Chenguang Zhu.
  2023.
\newblock Gpteval: Nlg evaluation using gpt-4 with better human alignment.
\newblock \emph{arXiv preprint arXiv:2303.16634}.

\bibitem[{Mahdi et~al.(2021)Mahdi, Dagli, Mustufa, and
  Nanivadekar}]{mahdi2021job}
Hussain~Falih Mahdi, Rishit Dagli, Ali Mustufa, and Sameer Nanivadekar. 2021.
\newblock Job descriptions keyword extraction using attention based deep
  learning models with bert.
\newblock In \emph{2021 3rd International Congress on Human-Computer
  Interaction, Optimization and Robotic Applications (HORA)}, pages 1--6. IEEE.

\bibitem[{OpenAI(2023)}]{openai2023gpt-4}
OpenAI. 2023.
\newblock Gpt-4 technical report.
\newblock \emph{arXiv preprint arXiv:2303.08774}.

\bibitem[{Ouyang et~al.(2022)Ouyang, Wu, Jiang, Almeida, Wainwright, Mishkin,
  Zhang, Agarwal, Slama, Ray et~al.}]{ouyang2022training}
Long Ouyang, Jeff Wu, Xu~Jiang, Diogo Almeida, Carroll~L Wainwright, Pamela
  Mishkin, Chong Zhang, Sandhini Agarwal, Katarina Slama, Alex Ray, et~al.
  2022.
\newblock Training language models to follow instructions with human feedback.
\newblock \emph{arXiv preprint arXiv:2203.02155}.

\bibitem[{Qin et~al.(2020)Qin, Zhu, Xu, Zhu, Ma, Chen, and
  Xiong}]{qin2020enhanced}
Chuan Qin, Hengshu Zhu, Tong Xu, Chen Zhu, Chao Ma, Enhong Chen, and Hui Xiong.
  2020.
\newblock An enhanced neural network approach to person-job fit in talent
  recruitment.
\newblock \emph{ACM Transactions on Information Systems (TOIS)}, 38(2):1--33.

\bibitem[{Singhal et~al.(2022)Singhal, Azizi, Tu, Mahdavi, Wei, Chung, Scales,
  Tanwani, Cole-Lewis, Pfohl et~al.}]{singhal2022large}
Karan Singhal, Shekoofeh Azizi, Tao Tu, S~Sara Mahdavi, Jason Wei, Hyung~Won
  Chung, Nathan Scales, Ajay Tanwani, Heather Cole-Lewis, Stephen Pfohl, et~al.
  2022.
\newblock Large language models encode clinical knowledge.
\newblock \emph{arXiv preprint arXiv:2212.13138}.

\bibitem[{Sun et~al.(2021)Sun, Zhuang, Zhu, Zhang, He, and
  Xiong}]{sun2021market}
Ying Sun, Fuzhen Zhuang, Hengshu Zhu, Qi~Zhang, Qing He, and Hui Xiong. 2021.
\newblock Market-oriented job skill valuation with cooperative composition
  neural network.
\newblock \emph{Nature communications}, 12(1):1992.

\bibitem[{Wang et~al.(2022)Wang, Li, Hirota, and Kandogan}]{wang2022machop}
Jin Wang, Yuliang Li, Wataru Hirota, and Eser Kandogan. 2022.
\newblock Machop: An end-to-end generalized entity matching framework.
\newblock In \emph{Proceedings of the Fifth International Workshop on
  Exploiting Artificial Intelligence Techniques for Data Management}, pages
  1--10.

\bibitem[{Wang et~al.(2021)Wang, Thompson, and Iyyer}]{wang2021phrase}
Shufan Wang, Laure Thompson, and Mohit Iyyer. 2021.
\newblock Phrase-bert: Improved phrase embeddings from bert with an application
  to corpus exploration.
\newblock \emph{arXiv preprint arXiv:2109.06304}.

\bibitem[{Wei et~al.(2022)Wei, Wang, Schuurmans, Bosma, Xia, Chi, Le, Zhou
  et~al.}]{wei2022chain}
Jason Wei, Xuezhi Wang, Dale Schuurmans, Maarten Bosma, Fei Xia, Ed~Chi, Quoc~V
  Le, Denny Zhou, et~al. 2022.
\newblock Chain-of-thought prompting elicits reasoning in large language
  models.
\newblock \emph{Advances in Neural Information Processing Systems},
  35:24824--24837.

\bibitem[{West et~al.(2021)West, Bhagavatula, Hessel, Hwang, Jiang, Bras, Lu,
  Welleck, and Choi}]{west2021symbolic}
Peter West, Chandra Bhagavatula, Jack Hessel, Jena~D Hwang, Liwei Jiang,
  Ronan~Le Bras, Ximing Lu, Sean Welleck, and Yejin Choi. 2021.
\newblock Symbolic knowledge distillation: from general language models to
  commonsense models.
\newblock \emph{arXiv preprint arXiv:2110.07178}.

\bibitem[{Xu et~al.(2016)Xu, Napoles, Pavlick, Chen, and
  Callison-Burch}]{xu2016optimizing}
Wei Xu, Courtney Napoles, Ellie Pavlick, Quanze Chen, and Chris Callison-Burch.
  2016.
\newblock Optimizing statistical machine translation for text simplification.
\newblock \emph{Transactions of the Association for Computational Linguistics},
  4:401--415.

\bibitem[{Zimmermann et~al.(2016)Zimmermann, Kotschenreuther, and
  Schmidt}]{zimmermann2016data}
Tim Zimmermann, Leo Kotschenreuther, and Karsten Schmidt. 2016.
\newblock Data-driven hr-r$\backslash$'esum$\backslash$'e analysis based on
  natural language processing and machine learning.
\newblock \emph{arXiv preprint arXiv:1606.05611}.

\end{thebibliography}

\appendix
\section{Details of Prompts}
We present the prompts utilized in G-eval for assessing consistency and factuality in Figures \ref{fig:cons} and \ref{fig:fact}, respectively. Additionally, the prompt used in our generation pipeline for generating triples of job descriptions, matched resumes, and unmatched resumes is provided in Figure \ref{fig:gen}.


\begin{figure*}[!ht]
    \begin{tcolorbox}[fontupper=\ttfamily, title={\small Consistency}]
    \scriptsize
    \# Instruction:\\
    As a hiring manager, your task is to evaluate job descriptions on a scale of 1-5. This scale represents the consistency of the job description, with 1 being completely inconsistent and 5 being fully consistent.
    Your evaluation should consider the alignment of job responsibilities, required skills, qualifications, and the overall tone of the job description, as well as the consistency between different sections of the job description. Please ensure you fully understand these instructions before proceeding.\\
    
    \# Evaluation Criteria:
    \begin{enumerate}[nolistsep]
        \item Completely Inconsistent: The job responsibilities, required skills, qualifications, and overall tone of the job description are not aligned. The description is confusing and does not provide a clear understanding of the job. Additionally, there are significant inconsistencies between different sections of the job description, making it confusing and unclear.
        \item Mostly Inconsistent: There are some elements of the job description that align, but there are significant inconsistencies between different sections that make the description unclear.
        \item Somewhat Consistent: The job description has a fair amount of alignment between the responsibilities, skills, and qualifications, as well as between different sections, but there are areas that could be improved for clarity.
        \item Mostly Consistent: The job description is mostly aligned, both within sections and between different sections, with only minor inconsistencies. The description provides a clear understanding of the job.
        \item Fully Consistent: The job responsibilities, required skills, qualifications, and overall tone of the job description are perfectly aligned. Additionally, there is a high level of consistency between different sections, resulting in a clear and comprehensive understanding of the job.
    \end{enumerate}
    ~\\
    \# Evaluation Steps:
    \begin{enumerate}[nolistsep]
        \item Carefully read the entire job description, focusing on the alignment between the job responsibilities, required skills, qualifications, and the overall tone of the description.
        \item Evaluate the overall consistency of the job description based on the provided criteria.
        \item Assign a consistency score ranging from 1 to 5, using the Evaluation Criteria as a guide.
    \end{enumerate}
    ~\\
    \# Required Skills:\\
    \{skills\}\\
    \\
    \# Required Experience:\\
    \{experience\}\\
    \\
    \# Job Description:\\
    \{job description\}\\
    \\
    \# Evaluation Form (scores ONLY):
    \end{tcolorbox}
    \caption{The prompt used for assessing the consistency in generated job descriptions. We use the same prompt, changing job descriptions to resumes, for resumes as well.}
    \label{fig:cons}
\end{figure*}

\begin{figure*}[!ht]
    \begin{tcolorbox}[fontupper=\ttfamily, title={\small Factuality}]
    \scriptsize
    \# Instruction:\\
    As a hiring manager, your task is to evaluate job descriptions on a scale of 1-5. This scale represents the factuality of the job description, with 1 being completely false and 5 being completely true. Your evaluation should consider the accuracy of the job responsibilities, required skills, qualifications, and the overall representation of the job role. Please ensure you fully understand these instructions before proceeding.\\
    
    \# Evaluation Criteria:
    \begin{enumerate}[nolistsep]
        \item Completely False: The job description does not match the job title at all. The responsibilities, required skills, and qualifications are misleading or incorrect.
        \item Mostly False: The job description has some elements of truth but contains significant inaccuracies or exaggerations in the responsibilities, required skills, or qualifications.
        \item Somewhat True: The job description is partially accurate. Some responsibilities, required skills, or qualifications may be overstated or understated.
        \item Mostly True: The job description is largely accurate, with minor discrepancies in the responsibilities, required skills, or qualifications.
        \item Completely True: The job description accurately represents the job title, responsibilities, required skills, and qualifications without any exaggeration or understatement.
    \end{enumerate}
    ~\\
    \# Evaluation Steps:
    \begin{enumerate}[nolistsep]
        \item Carefully read the entire job description, focusing on the job title, responsibilities, required skills, and qualifications.
        \item Evaluate the overall factuality of the job description based on the provided criteria.
        \item Assign a factuality score ranging from 1 to 5, using the Evaluation Criteria as a guide.
    \end{enumerate}
    ~\\
    \# Required Skills:\\
    \{skills\}\\
    \\
    \# Required Experience:\\
    \{experience\}\\
    \\
    \# Job Description:\\
    \{job description\}\\
    \\
    \# Evaluation Form (scores ONLY):
    \end{tcolorbox}
    \caption{The prompt used for assessing the factuality in generated job descriptions. We use the same prompt, changing job descriptions to resumes, for resumes as well.}
    \label{fig:fact}
\end{figure*}

\begin{figure*}[!ht]
    \begin{tcolorbox}[fontupper=\ttfamily]
    \scriptsize
    Write a job description for a ``\{job title\}'' job which require only skill set of ``\{list of skills\}'' and only previous job experience of ``\{list of experiences with augmented years for job description\}'' and a matching resume for a candidate with the name of ``\{sampled first name\}'' and having only skill set of ``\{list of skills\}'' and only previous job experience of ``\{list of experiences with augmented years for resume\}''. Then generate exactly the same resume (keeping everything the same) but excluding skill set of ``\{list of skills to be removed\}'' and ``\{the modification to last experience\}''. Don't include any extra skills and experience. But generate extra details about provided skills and job experience. The job description should only contain Job Title, Job Summary, Required Skills, and Responsibilities sections (only include few responsibilities). Resumes should only contain Personal Information (containing the provided first name and a matching generated last name and email), Education, Skills, and Experience sections. 
	The generated output should exactly be according the following structure:\\
    \\
	\#\#\#\#\#\# Job-description\\
    \\
	\#\# Job title\\
	.....\\
	\#\# Job Summary\\
	.....\\
	\#\# Required Skills\\
	.....\\
	\#\# Required Experience\\
	.....\\
	\#\# Responsibilities\\
	.....\\
    \\
	\#\#\#\#\#\# Resume 1\\
    \\
	\#\# Personal Information\\
	.....\\
	\#\# Education\\
	.....\\
	\#\# Skills\\
	.....\\
	\#\# Experience\\
	.....\\
    \\
	\#\#\#\#\#\# Resume 2\\
    \\
	\#\# Personal Information\\
	.....\\
	\#\# Education\\
	.....\\
	\#\# Skills\\
	.....\\
	\#\# Experience\\
	.....\\
    \\h
	output:
    \end{tcolorbox}
    \caption{The prompt used for generating documents in RJDB generation pipeline.}
    \label{fig:gen}
\end{figure*}

\end{document}